\PassOptionsToPackage{table}{xcolor}
\documentclass[lettersize,journal]{IEEEtran}
\usepackage{amsmath,amsfonts}
\usepackage{algorithm}
\usepackage{algorithmic}
\usepackage{array}
\usepackage[caption=false,font=normalsize,labelfont=sf,textfont=sf]{subfig}
\usepackage{textcomp}
\usepackage{stfloats}
\usepackage{url}
\usepackage{verbatim}
\usepackage{graphicx}
\usepackage{cite}
\usepackage{amsfonts}
\usepackage{amsmath}
\usepackage{tabularx}
\usepackage{booktabs}
\usepackage{multicol}
\usepackage{multirow}
\usepackage{overpic}
\usepackage[switch]{lineno}
\usepackage{xcolor}
\usepackage{arydshln}
\usepackage{bm}
\usepackage{makecell}
\usepackage{verbatim}
\usepackage{makecell}
\usepackage{multicol}
\usepackage{lipsum}
\usepackage{colortbl}
\usepackage{balance}

\usepackage{xspace}
\makeatletter
\DeclareRobustCommand\onedot{\futurelet\@let@token\@onedot}
\def\@onedot{\ifx\@let@token.\else.\null\fi\xspace}
\def\eg{\emph{e.g}\onedot} 
\def\ie{\emph{i.e}\onedot} 
 
\def\etc{\emph{etc}\onedot}

\makeatother

\definecolor{linkcolor}{RGB}{255,0,0}
\definecolor{urlcolor}{RGB}{255,105,180}
\definecolor{citecolor}{RGB}{0, 80, 200}
\definecolor{myorange}{RGB}{0,0,0}
\definecolor{mygray}{gray}{.93}
\definecolor{mygreen}{RGB}{78,172,91}
\definecolor{myred}{RGB}{235,50,35}
\usepackage{hyperref}
\hypersetup{colorlinks=true,linkcolor=linkcolor,urlcolor=urlcolor,citecolor=citecolor}

\begin{document}
\title{Purifying, Labeling, and Utilizing: A High-Quality Pipeline for Small Object Detection}
\author{Siwei Wang, Zhiwei Chen, Liujuan Cao, Rongrong Ji
\thanks{$^1$Our source code will be publicly released after acceptance.}
}

\markboth{Journal of \LaTeX\ Class Files,~Vol.~18, No.~9, September~2020}%
{How to Use the IEEEtran \LaTeX \ Templates}

\maketitle

\begin{abstract}
Small object detection is a broadly investigated research task and is commonly conceptualized as a ``pipeline-style'' engineering process.
In the upstream, images serve as raw materials for processing in the detection pipeline, where pre-trained models are employed to generate initial feature maps.
In the midstream, an assigner selects training positive and negative samples.
Subsequently, these samples and features are fed into the downstream for classification and regression.
Previous small object detection methods often focused on improving isolated stages of the pipeline, thereby neglecting holistic optimization and consequently constraining overall performance gains.
To address this issue, we have optimized three key aspects, namely \textbf{P}urifying, \textbf{L}abeling, and \textbf{U}tilizing, in this pipeline, proposing a high-quality \textbf{S}mall object detection framework termed PLUSNet.
Specifically, PLUSNet comprises three sequential components: the Hierarchical Feature Purifier~(HFP) for purifying upstream features, the Multiple Criteria Label Assignment~(MCLA) for improving the quality of midstream training samples, and the Frequency Decoupled Head~(FDHead) for more effectively exploiting information to accomplish downstream tasks.
The proposed PLUS modules are readily integrable into various object detectors, thus enhancing their detection capabilities in multi-scale scenarios.
Extensive experiments demonstrate the proposed PLUSNet consistently achieves significant and consistent improvements across multiple datasets for small object detection.$^1$
\end{abstract}

\begin{IEEEkeywords}
Small object detection, label assignment, frequency learning.
\end{IEEEkeywords}

\section{Introduction}
\label{introduction}
\IEEEPARstart{O}{bject} detection is a widely studied task that aims to locate and classify the objects of interest. 
As an integral extension of object detection, small object detection commands considerable importance, straddling the realms of theoretical inquiry and practical utility.
In general detection scenarios, the detection accuracy for small objects is often significantly lower compared to bigger objects.
Improving the performance on small objects is beneficial for overcoming detection bottleneck and greatly enhancing overall performance. 
In addition, in specific detection scenarios such as remote sensing~\cite{rabbi2020small, li2020cross, bashir2021small}, disaster rescue~\cite{zhang2022exploring, pi2020convolutional}, intelligent transportation~\cite{wu2020self, liu2021survey, chen2020survey}, medical applications\cite{hu2022small, elakkiya2022imaging}, the dominant presence is the abundance of small objects. 
In such cases, enhancing the performance of small object detection greatly contributes to progress in these fields.

Current small object detection pipeline, similar to a factory assembly line, can be divided into three stages: upstream, midstream, and downstream. 
In the upstream stage, the image goes through a pre-trained model to obtain feature maps, a process akin to feeding raw materials (\ie, images) into a machine (\ie, the pre-trained model) to produce initial parts (\ie, feature maps).
Subsequently, the feature maps along with candidate samples are fed into the midstream, where an assigner acts as a quality inspector to assess the samples, and assign positive or negative labels based on its evaluation. 
Finally, training samples and feature maps enter the downstream detection stage, where the detection head performs classification and regression tasks to produce results.

Nevertheless, across the three crucial stages of the pipeline, small object features are prone to loss, and low-quality small object features limit information utilization, leading to a significant degradation in the detection performance of small objects. 
Specifically, we identify three issues in widely-used small object detection frameworks:
\begin{figure}[t!]
\centering
\includegraphics[width=\linewidth]{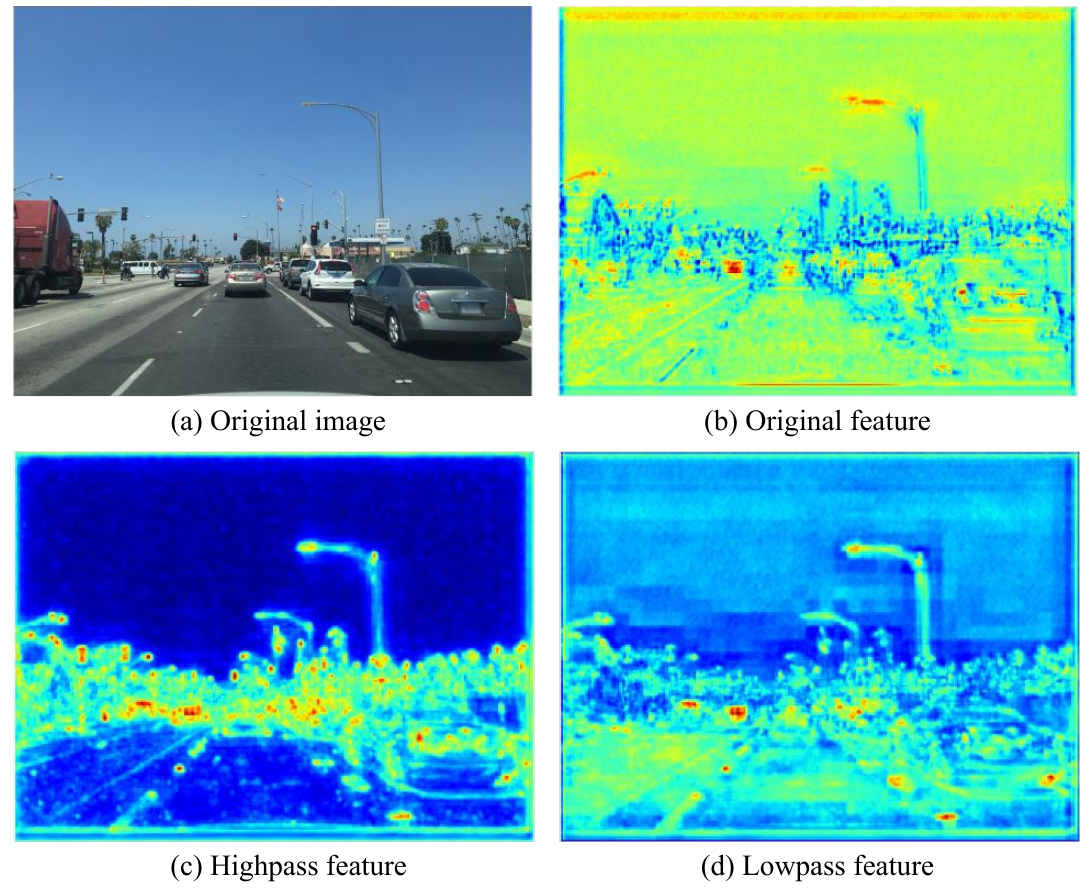}
\caption{Visualization of the lowest-level feature map. The original feature map contains a large amount of noise. After separating high and low frequency information through Fast Fourier Transform~\cite{nussbaumer1982fast}, highpass feature focus on small objects (\eg, traffic lights, vehicles, pedestrians), while lowpass feature concentrates more on overall semantic information (\ie, the street scene).}
\label{fig2}
\end{figure}
(1) The noisy low-frequency information in the bottom layers of the Feature Pyramid Network~(FPN)~\cite{lin2017feature} overwhelms the information of small objects.
As visualized in Fig.~\ref{fig2}(b), the lowest-level feature map originally designed for detecting small objects exhibits conspicuous noise.
We attribute this phenomenon to the effect of feature fusion in the FPN.
The top-down fusion manner continuously integrates the semantic information from the top-level with the detailed information, which inadvertently overshadows the information of small objects.
(2) The single-criterion label assignment greatly restricts the number of positive sample for small objects. 
Many mainstream detectors~\cite{ren2015faster, lin2017focal, tian2019fcos, carion2020end} depend solely on the widely used IoU-based label assignment. 
Unfortunately, the IoU~(Intersection over Union), fundamentally the Jaccard coefficient in mathematics~\cite{zhou2019iou,ayub2018jaccard}, possesses an inherent flaw: it is sensitive to the sizes of sets.
When a significant difference exists in the sizes of two sets, the Jaccard similarity is likely to lose accuracy, as it solely considers the ratio of intersection to union, ignoring the size discrepancy between sets.
This issue becomes particularly pronounced in small object detection, leading to various problems such as deviation and size issues, as illustrated in Fig.~\ref{fig1}(a).
(3) Current detection heads commonly share identical feature representations and structures for both classification and regression tasks, which is detrimental to small object detection. 
Small objects are inherently susceptible to interference from background and noise. 
For example, in Fig.~\ref{fig2}(a), the features of pedestrians may easily blend with those of background vehicles.
Directly performing classification and regression on such features simultaneously would confuse the detector's judgment.
Additionally, standard detection heads typically employ the same structure for distinct classification and regression.
Employing dedicated structures tailored to each task would constitute a more rational and effective design choice.

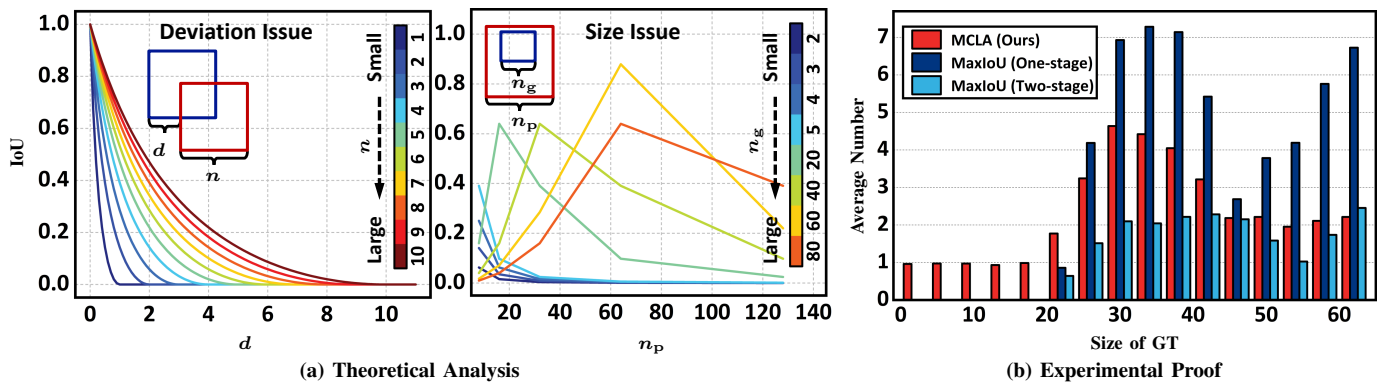
\begin{figure*}[!t]
\centering
\begin{overpic}[width=1\linewidth]{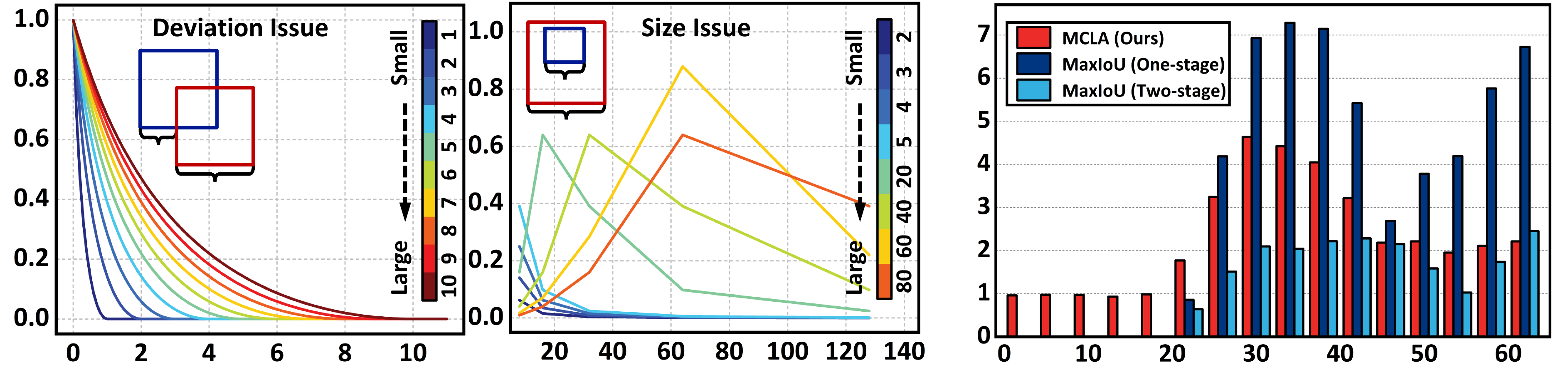}
	\put(-1.2, 11.8){\rotatebox{90}{\scriptsize\textbf{IoU}}}
	\put(60.4, 6.9){\rotatebox{90}{\scriptsize\textbf{Average Number}}}
	\put(9.2, 12.6){\scriptsize{\bm{$d$}}}
	\put(13.1, 10.8){\scriptsize{\bm{$n$}}}
	\put(24.5, 12.8){\rotatebox{90}{\scriptsize{\bm{$n$}}}}
	\put(15.5, -1.5){\scriptsize{\bm{$d$}}}
	\put(35.5, 17.6){\scriptsize{\bm{$n_{\mathrm{g}}$}}}
	\put(35.5, 14.8){\scriptsize{\bm{$n_{\mathrm{p}}$}}}
	\put(53.0, 12.8){\rotatebox{90}{\scriptsize{\bm{$n_{\mathrm{g}}$}}}}
	\put(44.8, -1.5){\scriptsize{\bm{$n_{\mathrm{p}}$}}}
	\put(78, -1.5){\scriptsize{\textbf{Size of GT}}}
	\put(20, -3.6){\footnotesize\textbf{(a) Theoretical Analysis}}
	\put(71.8, -3.6){\footnotesize\textbf{(b) Experimental Proof}}
\end{overpic}
\vspace{5pt}
\caption{ 
	Comparison of the single-criterion and multi-criteria label assignment in small object detection. (a) Deviation issue: Even when the proposal is perfectly aligned in size with the ground truth~(GT), minor deviations between them can lead to a rapid decrease in the IoU, especially for small objects. Size issue: Despite being fully contained within the GT, small-sized proposals struggle to achieve competitive IoU comparable to larger ones. (b) The distribution of positive sample numbers for mainstream detectors~\cite{lin2017focal, ren2015faster} and our PLUSNet. The representative MaxIoU methods completely ignore the small-sized objects, while our proposed MCLA assigns samples in a more reasonable manner. (Experimental details are available in Sec. \ref{exp_se}.)
}
\label{fig1}
\end{figure*}

To address the above three issues in a unified and effective manner, we propose PLUSNet, which consists of three mutually collaborative components as follows:
(1) To obtain informative yet noise-reduced feature representations, we propose Hierarchical Feature Purifier~(HFP) for feature purification at the upstream feature fusion stage. 
HFP introduces Fast Fourier Transform~(FFT)~\cite{nussbaumer1982fast} to decompose image features into distinct low-frequency and high-frequency domains.
As shown in Fig.~\ref{fig2}(c) and Fig.~\ref{fig2}(d), the low-frequency components capture semantic information (\ie, the street scene), while the high-frequency components eliminate the interference of background and noise, effectively highlighting small objects.
(2) To improve the overall quality of training samples, we introduce a novel Multiple Criteria Label Assignment~(MCLA) to increase the number of positive sample for small objects.
Specifically, MCLA is a multi-criteria paradigm that includes three metrics: the Position Offset Criterion~(POC), the Shape Constraint Criterion~(SCC) and the original IoU criterion.
MCLA effectively mitigates the issues in  Fig.~\ref{fig1}(a) at no cost.
As illustrated in Fig.~\ref{fig1}(b), by leveraging complementary criteria to the Jaccard coefficient, MCLA mitigates the inherent skewing issue posed by large objects, balances the average number of positive sample across different scales, and achieves equilibrium between the quality and diversity of positive samples.
(3) Finally, we further introduce a Frequency Decoupled Head~(FDHead) to intentionally bias attention towards relevant information while reducing interference from detrimental cues. 
For classification branch, FDHead focuses more on semantic information and utilizes fully connected layers, which is capable of establishing complex mapping relationships. 
For regression branch, FDHead emphasizes contour details and employs convolution layers instead, for the convolutions naturally excel in spatial awareness~\cite{wu2020rethinking}.
By leveraging the synergistic capabilities of these three PLUS modules,  the input image undergoes a more thorough feature purification, a more superior label assignment, as well as more specialized detection, culminating in the generation of accurate detection results.

The main contributions can be summarized as follows:
First, we take a holistic view and identify three critical issues in the small object detection pipeline. To address these challenges, we propose a novel PLUSNet, which jointly optimizes the pipeline at three critical stages.
Second, we alleviate the existing issues in small object detection from three angles: purifying, labeling, and utilizing. This enables the detector to possess cleaner features, an increased number of high-quality positive sample for small objects, and enhanced feature utilization capabilities.
Third, the three proposed modules are plug-and-play in detection pipeline, seamlessly integrated into the baseline algorithm, resulting in a significant and consistent improvement in detection performance, particularly for small objects. 
Extensive experiments conducted on multiple benchmark datasets, including SODA-D~\cite{cheng2023towards}, AI-TOD~\cite{wang2021tiny} and MS COCO~\cite{lin2014microsoft}), validate the effectiveness and generalizability of our proposed approach.

\section{Related Work}
\subsection{Small Object Detection}
Small object detection is a subset of object detection, and mainstream detectors strive to improve the detection performance of small objects in order to enhance multi-scale detection accuracy. 
For instance, anchor-based detectors~\cite{ren2015faster, lin2017focal, dai2021dynamic, zhang2020bridging} often employ a divide-and-conquer design using FPN to detect small objects by utilizing higher-resolution features from the lower layers.  
Anchor-free methods~\cite{tian2019fcos, law2018cornernet, zhou2019objects, yang2019reppoints}, on the other hand, adopt a dense prediction strategy by introducing more prediction points, thereby increasing the coverage and recall rate for small objects. 
In the case of query-based methods, the pioneering work DETR~\cite{carion2020end}, due to its lack of local perception capability, performs poorly in small object detection. 
Deformable-DETR~\cite{zhu2020deformable} improves the detection performance of small objects by incorporating multi-scale feature maps into its training process. 
Specifically, previous research in small object detection focused on acquiring more high-quality features for small objects through techniques like data manipulation~\cite{kisantal2019augmentation, zoph2020learning, yu2020scale}, super resolution~\cite{bai2018sod, noh2019better}, and feature fusion~\cite{gong2021effective,  hong2021sspnet}. 
Recently, researchers have explored the benefits of introducing Gaussian probability models in network learning to enhance small object detection performance.
For example, GWD~\cite{yang2021rethinking} and NWD~\cite{wang2021normalized} propose Gaussian distribution modeling for priors, while RFLA~\cite{xu2022rfla} designs a Gaussian-based label assignment method based on this modeling.
DCFL~\cite{xu2023dynamic} introduces a coarse-to-fine label assignment strategy, refining the selection process for high-quality positive samples. 
The recent CFINet~\cite{yuan2023small} achieves superior accuracy by employing a coarse-to-fine pipeline that ensures an adequate number of high-quality proposals. 
In summary, existing methods for small object detection often remain confined to improving a particular stage of detection pipeline, and the bottleneck effect still restricts overall performance improvement. 
Our approach, on the other hand, optimizes the entire pipeline, resulting in an overall enhancement of small object detection.

\subsection{Label Assignment}
Label assignment is crucial in training models, as it directly affects the distribution of training samples. 
The most common approach, MaxIoU label assignment~\cite{ren2015faster}, is widely used in both anchor-based and anchor-free detectors. 
It measures the quality of samples based on the IoU between proposals and ground truths, thus determining positive and negative samples. 
ATSS~\cite{zhang2020bridging} analyzes performance differences between anchor-based and anchor-free detectors, and then proposes a strategy to adaptively assign positive and negative samples. 
FSAF~\cite{zhu2019feature} introduces the concept of effective and ignored regions for objects, selecting positive and negative samples based on whether the center of the detection box falls within these regions. 
Furthermore, some improved label assignment methods aim to obtain better training samples from different perspectives.
OTA~\cite{ge2021ota} formulates label assignment as an optimal transport problem, optimizing transportation costs to achieve better assignment results. 
FreeAnchor~\cite{zhang2019freeanchor} abandons IoU-based design and uses maximum likelihood theory to select positive and negative samples.
Although various label assignment methods differ, they typically rely on a single sample evaluation metric, such as IoU, Wasserstein distance, or Hungarian cost. 
We have observed that this reliance on a single metric limits the diversity of training samples, particularly for small objects. 
To address this issue, we propose a multi-metric assignment strategy.

\subsection{Frequency Learning in Convolutional Neural Networks}
The frequency serves as an indicator of the intensity variations within the image, and frequency domain information greatly aids in image understanding. 
It is widely acknowledged that low-frequency information corresponds to the semantic content of an image, while high-frequency information corresponds to fine-grained details. 
The inspiring work~\cite{wang2020high} extensively investigates the impact of frequency on the robustness and accuracy of CNNs~(Convolutional Neural Networks), discovering that CNNs heavily rely on high-frequency information.
DCTNet~\cite{xu2020learning} provides a detailed analysis of the role of frequency domain in deep learning and proposes transforming images from the spatial domain to the frequency domain, enabling image compression with reduced communication bandwidth while achieving higher precision. 
IRN~\cite{xiao2020invertible} examines the main reason for the loss of information during image down-sampling can be attributed to the loss of high-frequency signals.
To address this, it cleverly employs wavelet transforms to preserve high-frequency information, enabling lossless and reversible image scaling. 
Numerous studies have already demonstrated the efficacy of frequency learning in improving the model's comprehension of images. 
However, the utilization of frequency domain to assist object detection has been largely unexplored. 
FDCOD~\cite{zhong2022detecting} pioneers the introduction of a frequency enhancement module in camouflage object detection, utilizing the DCT~(Discrete Cosine Transform) to extract frequency information. 
In our proposed approach, we employ the commonly used FFT and IFFT~(Inverse FFT) in frequency processing to aid in feature purification, thereby obtaining clean features of small objects. 
Moreover, the proposed FDHead also utilizes low-frequency and high-frequency components separately based on the distinct characteristics of the classification and regression tasks, effectively enhancing the performance of detection.

\section{Method}
\label{method}
\subsection{Overview} 
\begin{figure*}[t!]
\centering
\begin{overpic}[width=1\linewidth]{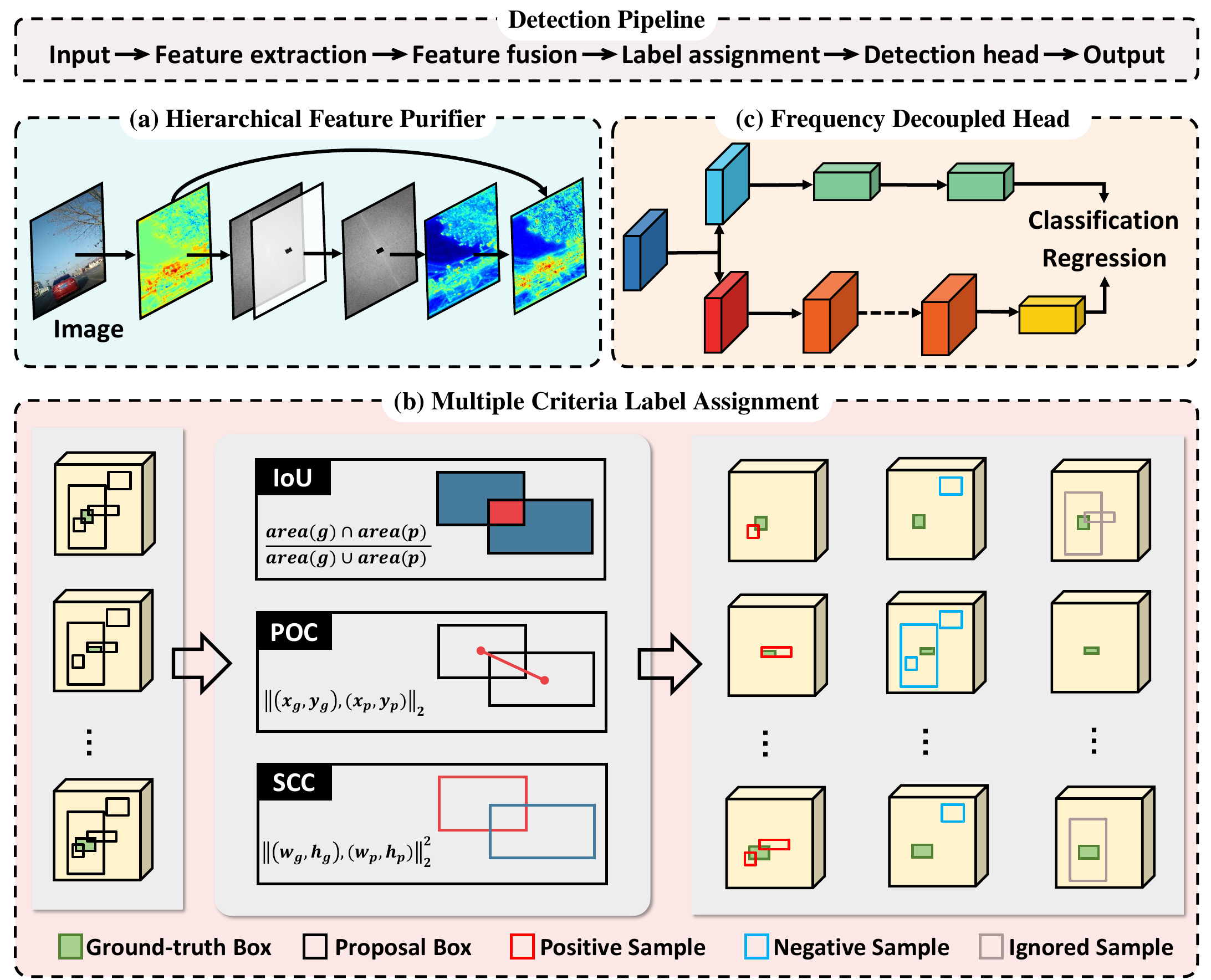}
	\put(13.7, 52.8){\scriptsize{\bm{$X_l$}}}
	\put(18.7, 52.8){\scriptsize{\bm{$S$}}}
	\put(23.3, 52.8){\scriptsize{\bm{$M$}}}
	\put(30.8, 52.8){\scriptsize{\bm{$S^\prime$}}}
	\put(37.2, 52.8){\scriptsize{\bm{$X_l^\prime$}}}
	\put(45.0, 52.8){\scriptsize{\bm{$\tilde{X}_{l}$}}}
	\put(53.0, 55.6){\scriptsize{\bm{$F_{roi}$}}}
	\put(55.5, 60.4){\scriptsize{\bm{$\mathtt{FFT}$}}}
	\put(61.3, 62.1){\scriptsize{\bm{$F_L$}}}
	\put(61.3, 51.7){\scriptsize{\bm{$F_H$}}}
	\put(79.0, 62.2){\scriptsize{\bm{$F_{cls}$}}}
	\put(84.5, 51.6){\scriptsize{\bm{$F_{reg}$}}}
	\put(81.0, 55.8){\scriptsize{\bm{$\mathtt{GAP}$}}}
	\put(62.1, 55.8){\scriptsize{\bm{$\mathtt{Conv}$}}}
	\put(71.5, 55.8){\scriptsize{\bm{$\mathtt{Conv}$}}}
	\put(63.5, 66.3){\scriptsize{\bm{$\mathtt{Fc}$}}}
	\put(74.4, 66.3){\scriptsize{\bm{$\mathtt{Fc}$}}}
\end{overpic}
\caption{Components of our proposed PLUSNet. We present the lowest-level feature in Hierarchical Feature Purifier, and it can be observed that the features of small objects (traffic-light, traffic-sign, \etc) become more distinct.}
\label{fig3-2}
\end{figure*}
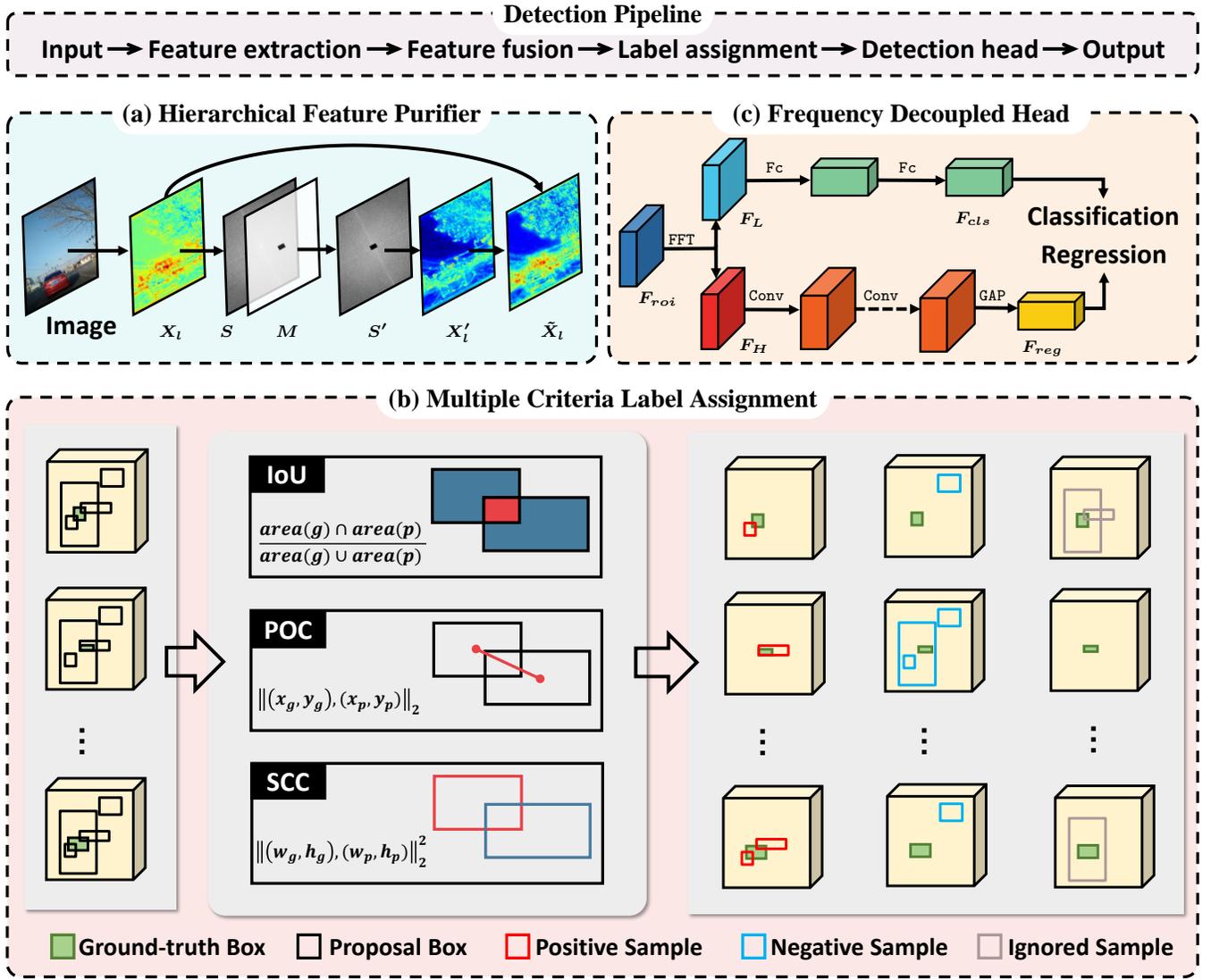

Over recent years, small object detection has evolved into a well-established paradigm. 
As illustrated in Fig.~\ref{fig3-2}, a typical detection pipeline generally encompasses six key components:
(1) Input of images and associated ground-truth annotations; 
(2) Image feature extraction, primarily performed by pre-trained networks like ResNet~\cite{he2016deep}, ResNeXt~\cite{xie2017aggregated}, HourglassNet~\cite{law2018cornernet}; 
(3) Feature fusion, which integrates semantic information with detailed information to enhance feature utilization, is commonly achieved using FPN~\cite{lin2017feature}; 
(4) Label assignment, which selects the positive and negative training samples; 
(5) Detection head, which is responsible for completing classification and regression tasks; 
(6) Output of results. 

Based on the pipeline paradigm, the feature maps provided by the upstream play a foundational role in subsequent detection.
The distribution of training samples in the midstream determine the generalization capability of the model, while the downstream is focused on the detection task, directly impacting the results of classification and regression. 
However, previous methods often focus solely on local aspects of the pipeline. 
To the best of our knowledge, we are the first to attempt a holistic optimization in small object detection.
Fig.~\ref{fig3-2} illustrates our proposed PLUS modules. 
We have made improvements at the upstream, midstream, and downstream stages of the pipeline, specifically in feature fusion, label assignment, and detection head. 
These complementary enhancements collectively improve the effectiveness of small object detection networks.

\subsection{Hierarchical Feature Purifier}
The majority of detection frameworks incorporate a FPN for image feature fusion.
However, the top-down fusion strategy employed by the FPN tends to propagate excessive high-level semantics into lower-resolution feature maps, which is detrimental to small object detection.
To address the issue of semantic information diluting fine-grained details in small object detection, we propose a Hierarchical Feature Purifier~(HFP). 
HFP innovatively applies Fourier Transform to 2D feature maps, accomplishing feature purification in the frequency domain. 

HFP is employed in low levels of FPN to filter out excessive semantic information while removing harmful noise, since small objects are primarily detected in these low-level features~\cite{gong2021effective, lin2017feature}.
As shown in Fig.~\ref{fig3-2}(a), we first select features $\mathbf{X}_{l} \in \mathbb{R}^{H \times W \times C}$ below the $r$-th level in FPN, since small objects are primarily detected in these low-level features. 
Subsequently, we adopt the Fast Fourier Transform~(FFT) to the selected features, transforming it into frequency spectrum maps $\mathbf{S} \in \mathbb{R}^{H \times W \times C}$. 
Then, a mask $\mathbf{M}\in \mathbb{R}^{H \times W}$ is applied to $\mathbf{S}$ for highpass filtering, thus generating the filtered maps $\mathbf{S}^{\prime} \in \mathbb{R}^{H \times W \times C}$. 
To control the degree of filtration, we introduce a hyper-parameter $\mu$ to remove low-frequency information while preserving clean details.
Considering lower-level features require stronger filtering, we design an adaptive filtering strength mask, described as follows:
\begin{align}
K = \mu*(\frac{r-l}{r}), 
\end{align}
\begin{align}
\mathbf{M}_{(i, j)} = \begin{cases}
	0, \text{if} \left| i - \frac{H}{2} \right| \leq \frac{K}{2}\ \text{and} \ \left| j - \frac{W}{2} \right| \leq \frac{K}{2}, \\
	1, \text{otherwise.} 
\end{cases}
\end{align}
where $l$ is the level of $\mathbf{X}_{l}$, in other words, the $l$-th level. And $\left| \cdot \right|$ denotes the operation of taking absolute value. 
Then, inverse FFT is performed on the filtered frequency spectrum maps $\mathbf{S}^{\prime} $ to obtain new feature maps $\mathbf{X}_{l}^{\prime}\in \mathbb{R}^{H \times W \times C}$ that emphasize refined focal details.
Finally, to mitigate potential loss of global perception, we utilize a residual structure to fuse the information from $\mathbf{X}_{l}$ and $\mathbf{X}_{l}^{\prime}$.
The purified feature map $\mathbf{\tilde{X}}_{l} \in \mathbb{R}^{H \times W \times C}$ is then obtained and fed into the downstream of the pipeline for precise classification and localization of small objects.
The process described above is formulated as follows:
\begin{equation}
\mathbf{\tilde{X}}_{l} = \mathtt{IFFT}\left (\mathbf{M} \odot \mathtt{FFT}(\mathbf{X}_{l}) \right ) * \omega + \mathbf{X}_{l}, 
\end{equation}
where $\omega$ is a weight parameter in residual structure, $l\in [0,r)$. $\odot$ denotes the element-wise production.

\subsection{Multiple Criteria Label Assignment}
As the upstream feature maps are confirmed, numerous predefined priors (anchors, points, \etc) are also generated within the feature maps, which then proceed to the midstream for evaluation by the assigner.
Prevalent label assignments often rely on a single IoU criterion, thereby inheriting the limitations of Jaccard coefficient into the label assignment process, which is especially pronounced in small object detection as aforementioned. 
It is imperative to adopt a multi-faceted approach for sample evaluation, considering diverse perspectives. 
Following this principle, we design a simple yet effective Multiple Criteria Label Assignment~(MCLA).

For the assigner, it receives a set of ground truth boxes, $\mathcal{B}_{g}$, and a set of proposals, $\mathcal{B}_{p}$. 
And then it evaluates the score for each proposal to determine their eligibility as positive samples. 
As depicted in the Alg.~\ref{alg1}, the MCLA evaluation process is as follows: 
First, MCLA obtains the center coordinates and dimensions of each box in $\mathcal{B}_{g}$, resulting in a four-element vector $(x_g, y_g, w_g, h_g)$.
And then, these vectors are continuously concatenated to form a matrix $\bm{\mathcal{M}}_g^{m \times 4}$, where $m$ is the number of ground-truth boxes. 
Similarly, MCLA constructs the proposal matrix $\bm{\mathcal{M}}_p^{n \times 4}$ using the same procedure, where $n$ is the number of proposals. 
Next, MCLA decouples the position and shape of the boxes, calculating their Mean Square Errors~(MSEs), $\bm{E}_1^{m \times n}$ and $\bm{E}_2^{m \times n}$. 
MSE is a complementary metric to the Jaccard coefficient in mathematics~\cite{ayub2018jaccard}, and we employ it to emphasize small-sized training samples. 
Subsequently, considering that $\bm{E}_1$ is sensitive to image size variance, we normalize it using min-max normalization to obtain $\bm{E}_1^{norm}$. 
With these preparations completed, MCLA proceeds to compute its three evaluation metrics between each ground-truth and proposal: $S_{IoU}$, $S_{POC}$, and $S_{POC}$. 
IoU is calculated using the original algorithm, while POC~(Position Offset Criterion) and SCC~(Shape Constraint Criterion) are transformed to a score distribution in the range $[0, 1]$ using non-linear mapping function, which also helps to mitigate the adverse effects of outliers. 
Finally, the three scores are weighted and normalized to obtain the final score, which comprehensively reflect the quality of each proposal. 
Moreover, due to the normalized weighting, the adjustment factors $20$ and $0.25$ in the non-linear mapping function have minimal impact on the results, as their influence can be attributed to the $\lambda$ weights. 
As is shown in Fig.~\ref{fig3-2}(b), the comprehensive MCLA enables a fair assessment of object scales, effectively mitigating the problem of imbalanced training sample scales, and providing a large number of high-quality training samples, which contributes to overall performance improvement.

\begin{algorithm} [t!]
\caption{Calculating Proposals' Score by MCLA}
\label{alg1}
\textbf{Input}: Ground-Truth box set $\mathcal{B}_{g}$, Proposal box set $\mathcal{B}_{p}$\\
\textbf{Output}: Score matrix {$\mathcal{S}^{m \times n}$}
\begin{algorithmic}[1]
	\STATE get $(x_g, y_g, w_g, h_g)$ of each ground-truth box from $\mathcal{B}_{g}$, and form the matrix $\bm{\mathcal{M}}_g^{m \times 4}$ 
	\STATE get $(x_p, y_p, w_p, h_p)$ of each proposal box from $\mathcal{B}_{p}$, and form the matrix $\bm{\mathcal{M}}_p^{n \times 4}$ 
	\STATE $\bm{E}_1^{m \times n} = \left\|\bm{\mathcal{M}}_g[:,:2], \bm{\mathcal{M}}_p[:,:2]\right\|_2$
	\STATE $\bm{E}_2^{m \times n} = {\left\|\bm{\mathcal{M}}_g[:,2:], \bm{\mathcal{M}}_p[:,2:]\right\|_2^2}$
	\STATE $\bm{E}_1^{norm} = (\bm{E}_1 - \min(\bm{E}_1)) / (\max(\bm{E}_1) - \min(\bm{E}_1))$
	\FORALL{ ${b}_{g} \in \mathcal{B}_{g} $, ${b}_{p} \in \mathcal{B}_{p} $ }
	\STATE calculate $S_{IoU}$ ($S_1$) according to original MaxIoU Assigner
	\STATE $S_2 = S_{POC} = (1 + \sqrt{20 * E_1^{norm}[b_g, b_p]})^{-1}$
	\STATE $S_3 = S_{SCC} = (1 + \sqrt{0.25 * E_2[b_g, b_p]})^{-1}$
	\STATE $\mathcal{S}[b_g, b_p] = \frac{1}{\lambda_1+\lambda_2+\lambda_3} \sum_{i=1}^{3} \lambda_i S_i$
	\ENDFOR
	\STATE  \textbf{return} $\mathcal{S}^{m \times n}$
\end{algorithmic}
\end{algorithm}

\subsection{Frequency Decoupled Head}
Once the purified features and meticulously curated samples are transferred to the downstream, the detection head's duty is to derive cues from the distinctive traits of these samples, seamlessly guiding them back to their rightful places, and accurately categorizing them. 
While classification and regression are two distinct tasks, it may be suboptimal for most detection heads to adopt identical structures and features.
To handle the distinctive focuses of the classification and regression tasks, and to reduce the interference of noisy information, we propose the Frequency Decoupled Head~(FDHead). 

The FDHead, as depicted in the Fig.~\ref{fig3-2}(c), incorporates separate branches for classification and regression:
(1) Starting from the RoIAlign-generated feature map $\bm{F_{roi}}\in \mathbb{R}^{h \times w \times c}$, we employ FFT filtering to extract low-frequency feature component $\bm{F_L}\in \mathbb{R}^{h \times w \times c}$ that encompass semantic information, and high-frequency feature component $\bm{F_H}\in \mathbb{R}^{h \times w \times c}$ that capture detailed contour information.
(2) In the classification branch, the low-frequency component $\bm{F_L}$ undergoes two fully connected layers, to obtain the feature vector $\bm{F_{cls}}\in \mathbb{R}^{1 \times 1 \times 1024}$ for classification.
(3) In the regression branch, the high-frequency component $\bm{F_H}$ is processed through a convolution layer for dimensionality expansion. 
Subsequently, several convolution blocks are applied, followed by the global average pooling (GAP) to obtain the feature vector $\bm{F_{reg}}\in \mathbb{R}^{1 \times 1 \times 1024}$ for regression. 
The process described above can be mathematically represented by the following formula. 
\begin{align}
&\bm{F_L}=\mathtt{BR_1}(\mathtt{FFT}(\bm{F_{roi}}; D_l)), \;\; \bm{F_{cls}} = \mathtt{Fc_2}(\mathtt{Fc_1}(\bm{F_L})), \\
&\bm{F_H}=\mathtt{BR_2}(\mathtt{FFT}(\bm{F_{roi}}; D_h)), \;\; \bm{F_{reg}} = \mathtt{GAP}(\mathtt{\sigma}(\bm{F_H}; W)), 
\end{align}
where $\mathtt{BR}$ is the BN-RELU layer. $D_l$ and $D_h$ are cutoff frequency, which indicate the intensities of lowpass filtering and highpass filtering. $\mathtt{\sigma}$ represents the convolution blocks, and $W$ is the learnable parameters within the convolution layers.
Lastly, $\bm{F_{cls}}$ and $\bm{F_{reg}}$ pass through separate linear layers to obtain the classification and regression results. 
The model parameters are then updated through back propagation using cross-entropy loss for classification and Smooth-L1 loss for regression.

\begin{table*}[t!]
\centering
\caption{Comparison with state-of-the-art detection approaches on the SODA-D \textit{test-set}, where Faster RCNN is the baseline method.}
\resizebox{1.0\textwidth}{!}{
	\begin{tabular}{|c|c|c|ccc|cccc|}
		\hline
		Method & Backbone & Schedule & $AP$  & $AP_{50}$  & $AP_{75}$  & $AP_{eS}$  & $AP_{rS}$  & $AP_{gS}$  & $AP_{N}$ \\
		\hline
		\hline
		\textit{\textbf{Query-based}} & \multicolumn{1}{c}{} & \multicolumn{1}{c}{} &       &       & \multicolumn{1}{c}{} &       &       &       &  \\
		\hline
		Deformable-DETR \cite{zhu2020deformable} & ResNet-50 & 50e   & 19.2  & 44.8  & 13.7  & 6.3  & 15.4  & 24.9  & 34.2  \\
		Sparse RCNN \cite{sun2021sparse} & ResNet-50 & 1$\times$ &  24.2 & 50.3 & 20.3 & 8.8 & 20.4 & 30.2 & 39.4  \\
		\hline
		\textit{\textbf{Anchor-free}} & \multicolumn{1}{c}{} & \multicolumn{1}{c}{} &       &       & \multicolumn{1}{c}{} &       &       &       &  \\
		\hline
		FCOS \cite{tian2019fcos} & ResNet-50 & 1$\times$    & 23.9  & 49.5  & 19.9    & 6.9  & 19.4  & 30.9  & 40.9  \\
		CornerNet \cite{law2018cornernet} & Hourglass-104 & 2$\times$    & 24.6 & 49.5 & 21.7 & 6.5 & 20.5 & 32.2 & 43.8 \\
		CenterNet \cite{zhou2019objects} & ResNet-50 & 70e    & 21.5  & 48.8  & 15.6  & 5.1  & 16.2  & 29.6  & 42.4 \\
		RepPoints \cite{yang2019reppoints} & ResNet-50 & 1$\times$    & 28.0  & 55.6  & 24.7  & 10.1 & 23.8  & 35.1  & 45.3  \\
		\hline
		\textit{\textbf{Anchor-based}} & \multicolumn{1}{c}{} & \multicolumn{1}{c}{} &       &       & \multicolumn{1}{c}{} &       &       &       &  \\
		\hline
		RetinaNet \cite{lin2017focal} & ResNet-50 & 1$\times$    & 28.2  & 57.6 & 23.7 & 11.9    & 25.2  & 34.1  & 44.2 \\
		Cascade RPN \cite{vu2019cascade} & ResNet-50 & 1$\times$ & 29.1 & 56.5 & 25.9 & 12.5 & 25.5 & 35.4 & 44.7 \\
		ATSS \cite{zhang2020bridging} & ResNet-50 & 1$\times$    & 26.8 & 55.6 & 22.1 & 11.7 & 23.9 & 32.2 & 41.3   \\
		YOLOX \cite{ge2021yolox} & CSP-Darknet & 70e & 26.7  & 53.4  & 23.0  & 13.6  & 25.1  & 30.9  & 30.4 \\
		DyHead \cite{dai2021dynamic} & ResNet-50 & 1$\times$ & 27.5  & 56.1  & 23.2  & 12.4  & 24.4  & 33.0  & 41.9 \\
		RFLA \cite{xu2022rfla} & ResNet-50 & 1$\times$    & 29.7  & 60.2  & 25.2  & 13.2  & 26.9  & 35.4  & 44.6 \\
		CFINet \cite{yuan2023small} & ResNet-50 & 1$\times$ & 30.7 & 60.8 & 26.7 & 14.7 & 27.8 & 36.4 & 44.6 \\
		\hline
		\textit{\textbf{Our method}} & \multicolumn{1}{c}{} & \multicolumn{1}{c}{} &       &       & \multicolumn{1}{c}{} &       &       &       &  \\
		\hline
		Faster RCNN \cite{ren2015faster} & ResNet-50 & 1$\times$    & 28.9  & 59.4  & 24.1  & 13.8  & 25.7  & 34.5  & 43.0 \\
		\rowcolor{gray!30}
		PLUSNet (ours) & ResNet-50 & 1$\times$ & 32.0 & 61.7 & 28.7 & 15.4 & 28.5 & 38.2 & 47.0 \\
        \rowcolor{gray!30}
        PLUSNet (ours) & ResNeXt-101~\cite{xie2017aggregated} & 1$\times$ & \textbf{32.5} & \textbf{62.3} & \textbf{28.9} & \textbf{16.1} & \textbf{28.8} & \textbf{38.5} & \textbf{48.1} \\
		\hline
	\end{tabular}
}
\label{tab:SODAResuts}
\end{table*}%

\section{Experiment}
\label{experiment}
\subsection{Setup}
\textbf{Dataset.}
To validate the effectiveness of our method on small object detection tasks, we conducted comparative experiments on two representative small object benchmark datasets, specifically SODA-D~\cite{cheng2023towards} and AI-TOD~\cite{wang2021tiny}.

SODA-D is the most recent benchmark dataset for small object detection, including $24,828$ high-quality images captured from autonomous driving scenarios, with a total of $278,433$ annotated small-sized instances. The average size of them is merely $20.31$ pixels.
AI-TOD consists of $28,036$ aerial and drone-viewed images with $700,621$ object instances across $8$ categories.  
It is a highly representative small object detection dataset in remote sensing, as the average size of objects is approximately $12.8$ pixels, significantly smaller than that of other aerial datasets.

\textbf{Evaluation Metrics.}
The detected objects in SODA-D are further classified into four degrees: $AP_{eS}$ (0 to 12 pixels), $AP_{rS}$ (12 to 20 pixels), $AP_{gS}$ (20 to 32 pixels), and $AP_{N}$ (32 to 45 pixels), quantitatively reflecting the detection performance of small objects.
Similar to SODA-D, AI-TOD further refines the detection metrics based on the size of objects into $AP_{vt}$ (2 to 8 pixels), $AP_{t}$ (8 to 16 pixels), $AP_{s}$ (16 to 32 pixels), and $AP_{m}$ (32 to 64 pixels).

\textbf{Experiment Settings.}
All experiments follow the $1\times$ training schedule of $12$ epochs, with implementation details aligned with the settings of the baseline algorithms. 
Specifically, all experiments are conducted using a single RTX3090 GPU.
The batch size is set to 4 for SODA-D, and 2 for AI-TOD.
The model uses the SGD (Stochastic Gradient Descent) optimization algorithm. 
The learning rate starts at $0.01$ and is coupled with a momentum of $0.9$, decaying at the 8th and 11th epochs.
The weight decay is set to $0.0001$.

\subsection{Comparision with State-of-the-art Methods}
\textbf{Results on SODA-D.} We conduct a comparison of PLUSNet with $14$ representative methods on the SODA-D benchmark.
As shown in Tab.~\ref{tab:SODAResuts}, query-based and anchor-free methods significantly lag behind anchor-based methods, particularly for extremely small and relatively small sizes. 
This indicates that there is still significant room for improvement in these methods for small object detection tasks. 
Among the anchor-based methods, RFLA~\cite{xu2022rfla} and CFINet~\cite{yuan2023small} are considered the state-of-the-art approaches in small object detection. 
By contrast, our proposed PLUSNet, utilizing only the basic Faster RCNN as the baseline, achieves a direct improvement of over $3$ $AP$ in detection accuracy, higher than state-of-the-art methods. 
Using ResNeXt-101 as the backbone network, our method achieves superior performance.
Furthermore, our approach surpasses the current best algorithms in all metrics, demonstrating its superiority.

\begin{table*}[t!]
\centering
\caption{Comparison results on AI-TOD \textit{test-set}. PLUSNet is based on Faster RCNN, and `$^+$' denotes using our proposed PLUS modules.}
\resizebox{1.0\textwidth}{!}{
	\begin{tabular}{|c|c|c|ccc|cccc|}
		\hline
		Method & Backbone & Schedule & $AP$  & $AP_{50}$  & $AP_{75}$  & $AP_{vt}$  & $AP_{t}$  & $AP_{s}$  & $AP_{m}$ \\
		\hline
		\hline
		RetinaNet~\cite{lin2017focal}   & ResNet-50	& 1$\times$ & 8.7	& 22.3 & 4.8 & 2.4 & 8.9 & 12.2 & 16.0     \\
		FCOS~\cite{tian2019fcos}    & ResNet-50 & 1$\times$ 	& 12.6	& 30.4 & 8.1 & 2.3 & 12.2 & 17.2 &  25.0 \\
		OTA~\cite{ge2021ota}  & ResNet-50 & 1$\times$ 	& 10.4	& 24.3  & 7.2 &  2.5 & 11.9 & 15.7 & 20.9   \\
		AutoAssign~\cite{zhu2020autoassign}  & ResNet-50 & 1$\times$ 	& 12.2	& 32.0  & 6.8 &  3.4 & 13.7 & 16.0 & 19.1   \\
		M-CenterNet~\cite{wang2021tiny} & DLA-34 & 1$\times$ & 14.5 & 40.7 & 6.4 & 6.1 & 15.0 & 19.4 & 20.4 \\ 
		ATSS~\cite{zhang2020bridging}  & ResNet-50 	& 1$\times$ & 12.8	& 30.6 & 8.5 & 1.9 & 11.6 & 19.5 & 29.2     \\
		Cascade RPN~\cite{vu2019cascade} & ResNet-50	& 1$\times$ 	& 13.3	& 33.5  & 7.8 & 3.9 & 12.9 & 18.1 & 26.3 \\
		Cascade RCNN~\cite{cai2018cascade}  & ResNet-50 & 1$\times$	 & 13.8	& 30.8 & 10.5 & 0.0 & 10.6 & 25.5 & 36.6 \\
		DotD~\cite{xu2021dot} & ResNet-50& 1$\times$ & 	16.1 & 39.2  & 10.6 & 8.3 & 17.6 & 18.1 & 22.1     \\
		Deformable-DETR~\cite{zhu2020deformable}& ResNet-50& 50e & 18.9 & 50.0  & 10.5 & 6.5 & 17.6 & 25.3 & 34.4    \\
		DAB-DETR~\cite{liu2022dab} & ResNet-50& 50e & 	22.4 & 55.6  & 14.3 & 9.0 & 21.7 & 28.3 & \textbf{38.7}    \\
		DetectoRS w/ NWD~\cite{wang2021normalized} & ResNet-50& 1$\times$ & 	20.8 & 49.3  & 14.3 & 6.4 & 19.7 & 29.6 & 38.3    \\
		DetectoRS w/ RFLA~\cite{xu2022rfla} & ResNet-50 & 1$\times$	 & 24.8	&55.2 & 18.5 & 9.3 & 24.8 & \textbf{30.3} & 38.2 \\
		\hline
		\textit{\textbf{Our method}} & \multicolumn{1}{c}{} & \multicolumn{1}{c}{} &       &       & \multicolumn{1}{c}{} &       &       &       &  \\
		\hline
		Faster RCNN~\cite{ren2015faster} & ResNet-50 & 1$\times$	& 11.1	& 26.3 & 7.6 & 0.0 & 7.2 & 23.3 & 33.6  \\
		DetectoRS~\cite{qiao2021detectors}  & ResNet-50 	& 1$\times$ 	& 14.8	& 32.8 & 11.4 & 0.0 & 10.8 & 28.3 & 38.0  \\
		\rowcolor{gray!30}
		PLUSNet (ours)   & ResNet-50 	& 1$\times$ 	& 21.7	& 53.4 & 13.6 & 8.3 & 21.0 & 27.6 & 36.5  \\
		\rowcolor{gray!30}
		DetectoRS$^+$ (ours) & ResNet-50 & 1$\times$	& \textbf{25.2}	& \textbf{55.6} & \textbf{18.8} & \textbf{10.6} & \textbf{25.3} & 29.9 & 38.2  \\
		\hline
	\end{tabular}
}
\vspace{-5pt}
\label{tab:AITODResuts}%
\end{table*}%
\textbf{Results on AI-TOD.} In the task of small object detection for remote sensing and drone scenes, our PLUSNet consistently achieves noteworthy performance advancements.
As shown in Tab.~\ref{tab:AITODResuts}, building upon the classic Faster RCNN, our approach incorporates optimized purification, labeling, and utilization strategies, yielding an improvement of over $10$ $AP$ in detection performance.
By applying the PLUS modules to the advanced DetectoRS, our method still achieves significant performance improvements, surpassing all approaches.

\subsection{Ablation Experiments}
\textbf{Ablation Study of Individual Component.} We conducted ablation study for the effectiveness of each proposed PLUS module, and the results are reported in Tab.~\ref{tab:ab1}. 
HFP introduces cleaner feature maps, thereby improving overall detection accuracy  from $28.9$ $AP$ to $30.0$ $AP$. 
However, the highpass filtering removes a significant amount of low-frequency information, leading to a decrease in the overall feature coupling. 
Due to the lack of sufficient small object samples and the absence of rational utilization of limited features, the detection accuracy for extremely small objects is adversely affected. 
With the application of MCLA, small-sized positive samples receive more attention, resulting in a substantial improvement from $12.6$ $AP_{eS}$ to $13.8$ $AP_{eS}$. 
Finally, FDHead enables better utilization of features and training samples, comprehensively enhancing the quality of the multi-scale detection pipeline, reaching $32.0$ $AP$ ultimately.
Through the combined effect of these three PLUS components, the performance of small object detection experiences systematic and significant improvement.

\textbf{Ablation Study in HFP.} 
(1) The divide-and-conquer design in  FPN allows its low levels to capture detailed information for detecting small objects, while the high levels primarily capture semantic information for detecting large objects. 
Our proposed HFP is specifically designed to eliminate redundant semantic information and enhance the representation of low-level features for small objects. 
To achieve this, we introduce a relay level, denoted as $r$, to control the hierarchical depth. 
The different impacts brought by various $r$ values are clearly observed in the Tab.~\ref{tab:supp_ab1}. 
When $r=0$, indicating the absence of the HFP component, the overall detection performance is only $31.4$ $AP$. 
With the introduction of HFP, significant improvements in detection accuracy are achieved, especially when $r=2$, where the $AP$ for small objects increases from $13.7$ to $15.4$, resulting in notable enhancement in small object detection capability and overall performance. 
However, when $r=4$, implying the absence of hierarchical strategy, all FPN layers undergo low-frequency feature filtering, leading to insufficient semantic information for detecting larger objects and resulting in a decrease in detection performance to $31.6$ $AP$.
The above results conclusively demonstrate the necessity and effectiveness of the hierarchical strategy in HFP.
(2) In HFP, low-frequency filtering is achieved by multiplying the mask $\mathbf{M}$ with the spectrum map $\mathbf{S}$, where the filtering strength is determined by the hyperparameter $\mu$. 
As $\mu$ increases, the filtering strength intensifies, resulting in the removal of more low-frequency information and enhancing the prominence of high-frequency details in the features, thus improving the detection capability for small objects. 
To explore the optimal filtering strength, a series of ablation experiments are conducted. 
As is shown in Tab.~\ref{tab:supp_ab2}, it can be observed that appropriately increasing the value of $\mu$ leads to improvements in $AP_{eS}$ and $AP_{rS}$. 
However, excessively high $\mu$ values result in significant loss of semantic information, leading to a decline in the detection capabilities of larger objects, as indicated by $AP_{gS}$ and $AP_{N}$. 
The best performance is achieved when $\mu$ is set to $0.05$. 
Additionally, we introduce residual structure to prevent the loss of global perception ability and use hyperparameter $\omega$ to control the weight of the purified features. 
Experimental results demonstrate that with $\mu$ fixed at the optimal value of $0.05$, setting $\omega$ to $0.3$ achieves the best detection performance.

\begin{table*}[ht]
    \centering
    \begin{minipage}{.45\linewidth}
        \setlength{\tabcolsep}{0.6mm}
        \caption{Analysis of each PLUS component on SODA-D \textit{test-set}.}
	\begin{tabular}{|c|ccc|ccccc|}
		\hline
		Baseline & HFP  & MCLA & FDHead & $AP$  & $AP_{eS}$  & $AP_{rS}$  & $AP_{gS}$  & $AP_{N}$ \\ 
		\hline
		\checkmark &  &  &  & 28.9 	&13.8 &	25.7 &	34.5 &	43.0\\
		\checkmark & \checkmark &  &  & 30.0 &	12.6 &	26.9 &	36.3 &	45.9 \\ 
		\checkmark &  \checkmark & \checkmark & & 30.6 &	13.8 &	27.1 &	36.7 &	45.6  \\
		\checkmark & \checkmark & \checkmark & \checkmark & \textbf{32.0}   &\textbf{15.4} &	\textbf{28.5} &	\textbf{38.2} &	\textbf{47.0}  \\
		\hline
	\end{tabular}
	\label{tab:ab1}

        \vspace{8pt}

        \setlength{\tabcolsep}{2.4mm}
        \caption{Results on SODA-D \textit{test-set} of filtering intensity $\mu$ and purified feature weight $\omega$.}
        \begin{tabular}{|cc|ccccc|}
            \hline
            $\mu$ & $\omega$ & $AP$  & $AP_{eS}$ & $AP_{rS}$ & $AP_{gS}$ & $AP_{N}$ \\
            \hline
            0.05 & 0.3 &  \textbf{32.0}& \textbf{15.4}&\textbf{28.5}&\textbf{38.2} &\textbf{47.0}\\ 
		  \hdashline[3pt/ 1.5pt]
		  0.02 & 0.3 &  31.8 & 14.5 &	28.2&38.2 &47.0 \\
		  0.10 &  0.3 & 31.7&15.0 &	28.5 &37.7 &46.7 \\
		  0.20& 0.3 & 31.8  &15.2 &28.5 &37.9 &46.7\\
		  \hline
		  0.05 & 0.1 &  31.8 &	15.0 &	28.2&38.2 &47.1 \\
		  0.05 & 0.2 &  31.8& 14.9&28.4&38.0 &46.9\\ 
		  0.05 &  0.4 & 31.6&14.7 &	28.1 &37.8 &46.9 \\
		  0.05 &  0.5 & 31.7&14.9 &28.3	 &37.8 &47.1 \\
		\hline
        \end{tabular}
        \label{tab:supp_ab2}
    \end{minipage}%
    \begin{minipage}{.45\linewidth}
        \setlength{\tabcolsep}{3.2mm}
        \caption{Results of different relay levels $r$ on SODA-D \textit{test-set}.}
        \begin{tabular}{|c|ccccc|}
            \hline
            $r$ & $AP$ & $AP_{eS}$ & $AP_{rS}$ & $AP_{gS}$ & $AP_{N}$ \\
            \hline
            0 & 31.4  & 13.7 & 27.9 & 38.0 & \textbf{47.3} \\
            1 & 31.8 & 14.7 & 28.2 & 38.0 & 46.9 \\
            2 & \textbf{32.0}  & \textbf{15.4} & \textbf{28.5} & \textbf{38.2} & 47.0 \\
            3 & 31.9 & 15.3 & 28.4 & 38.0 & 46.7 \\
            4 & 31.6 & 15.3 & 28.1 & 37.8 & 46.6 \\
            \hline
        \end{tabular}
        \label{tab:supp_ab1}

        \vspace{8pt}

        \caption {Ablation results of different score weights on SODA-D \textit{test-set}. With the introduction of normalization, we can fix the $\lambda_1$ to $1.0$ and only adjust the two hyper-parameters, $\lambda_2$ and $\lambda_3$.}
        \footnotesize
        \setlength{\tabcolsep}{1.8mm}{
	\begin{tabular}{|ccc|ccccc|}
		\hline
		$\lambda_1$ & $\lambda_2$ & $\lambda_3$&$AP$ & $AP_{eS}$  & $AP_{rS}$  & $AP_{gS}$  & $AP_{N}$ \\ 
		\hline
		1.0&2.5 & 1.0 &  31.6& 15.0 &28.1&37.7&46.2 \\
		1.0&3.5 & 1.0 &  31.9& 14.8&28.4&38.0 &47.0\\ 
		1.0&3.0 & 1.5 & 31.4 &14.3&27.9&37.7&46.6 \\
		1.0&3.0& 0.5 & 31.6& 14.7&28.5&37.5&46.5 \\
		\hline
		1.0&3.0& 1.0&\textbf{32.0}  &\textbf{15.4}&\textbf{28.5} &\textbf{38.2} &	\textbf{47.0} \\
		\hline
	\end{tabular}
	\label{tab:supp_ab3}}
        
    \end{minipage}
\end{table*}

\begin{table*}[ht]
    \centering
    \begin{minipage}{.45\linewidth}
        \setlength{\tabcolsep}{2.4mm}
        \caption{Effect of different FFT cutoff frequencies $D_h$ and $D_l$.}
	\begin{tabular}{|cc|ccccc|}
		\hline
		$D_l$ & $D_h$ &$AP$   & $AP_{eS}$  & $AP_{rS}$  & $AP_{gS}$  & $AP_{N}$ \\ 
		\hline
		0.80 & 0.10 &  31.9 &	14.7 &	28.5&38.1 &47.0 \\
		0.90 & 0.10 &  31.9 &	\textbf{15.7} &	\textbf{28.7} &37.9 &46.8\\ 
		0.85 &  0.05 & 31.9 &15.1 &	28.5 &38.0 &46.9 \\
		0.85& 0.15 & 31.9 &14.7 &28.6 &37.9 &47.0 \\
		\hline
		0.85& 0.10&\textbf{32.0}  &15.4&28.5 &\textbf{38.2}&	\textbf{47.0} \\
		\hline
	\end{tabular}
	\label{tab:supp_ab4}
    \end{minipage}%
    \begin{minipage}{.45\linewidth}
        \setlength{\tabcolsep}{2.2mm}
        \caption {Ablation results of using different frequency components in the classification and regression branch.  }
	\begin{tabular}{|cc|ccccc|}
		\hline
		$cls$ &$reg$&$AP$  & $AP_{eS}$  & $AP_{rS}$  & $AP_{gS}$  & $AP_{N}$ \\ 
		\hline
		$F_L$&$F_H$&  \textbf{32.0}& \textbf{15.4}&28.5&\textbf{38.2} &\textbf{47.0} \\
		\hdashline[3pt/ 1.5pt]
		$F_L$&$F_L$&  31.6& 15.0&28.4&37.4 &46.3\\ 
		$F_H$&$F_L$& 31.4 &14.5&28.1&37.5&46.0 \\
		$F_H$&$F_H$& 31.8& 15.1&\textbf{28.6}&38.0&46.5 \\
		\hline
	\end{tabular}
	\label{tab:supp_ab5}
    \end{minipage}
\end{table*}

\textbf{Ablation Study in MCLA.} In the MCLA, multiple criteria are combined in a weighted manner and then normalized to restrict the score between 0 and 1. 
The adjustment of weights essentially involves finding a balance between the Jaccard-based score and the MSE-based score. 
When these scores are finely tuned to reach a complementary state, the quality of training samples improves. 
From the results presented in Tab.~\ref{tab:supp_ab3}, it can be observed that the best performance is achieved when $\lambda_1$, $\lambda_2$ and $\lambda_3$ are set to $1.0$, $3.0$, $1.0$, respectively.

\textbf{Ablation Study in FDHead.} 
(1) The effectiveness of highpass filtering in FDHead is determined by the cutoff frequency $D_h$, which controls the preservation of high-frequency information. 
Higher $D_h$ values result in a reduction of retained low-frequency information, leading to more prominent high-frequency features. 
Conversely, the $D_l$ behaves in the opposite manner. 
Adjusting $D_l$ and $D_h$ directly impacts the redundancy and comprehensiveness of the feature. 
Analyzing the results presented in Tab.~\ref{tab:supp_ab4}, we find that FDHead exhibits robustness to fluctuations of cutoff frequency, as the detection outcomes remain relatively consistent within a specific range of cutoff values.
(2) In order to comprehensively demonstrate the rationality of the feature decoupling approach employed in our proposed FDHead, we conduct ablation experiments to examine the impact of utilizing different frequency components for classification and regression. 
As we all know, low-frequency signals in features contain rich semantic information, while high-frequency signals depict object contours. 
The decoupling of high and low-frequency features in FDHead aligns with this prior understanding, as further confirmed by the Tab.~\ref{tab:supp_ab5}. 
We can observe that utilizing low-frequency feature components for classification and high-frequency feature components for regression indeed achieves the best performance with $32.0$ $AP$. 
Conversely, when regressing with low-frequency semantic information and categorizing with high-frequency detail information, the performance declines to $31.4$ $AP$.

\subsection{Other Experiments}
\textbf{Generalization Experiment on MS COCO. }
\label{exp_ge}
To verify the generality of our PLUS modules across multi-scale object detection, we conduct extensive experiments on the widely-used MS COCO dataset~\cite{lin2014microsoft}, demonstrating that our method extends beyond small object detection.
In particular, we select several classic detectors and integrate the PLUS modules into them. 
As observed from the results in Tab.~\ref{tab:COCOResults}, we encouragingly find that our approach not only demonstrates plug-and-play capability, allowing it to be seamlessly integrated into various detectors, but also benefits from the robustness of the PLUS components for multi-scale detection. 
Through hierarchical strategies, weighted criteria, and frequency filtering, \etc, our high-quality detection pipeline consistently delivers significant performance improvements, even in datasets where small objects are not the primary focus, thus showcasing the consistent and remarkable performance enhancement brought forth by the high-quality pipeline.

\begin{table*}[thbp]
\centering
\footnotesize
\setlength{\tabcolsep}{4.0mm}{
	\small
	\caption{Results on COCO \textit{mini-val} set. The baseline results come from MMdetection~\cite{mmdetection}. `$^+$' denotes using our proposed PLUS modules.}
	\begin{tabular}{|c|c|ccc|ccc|}
		\hline
		Methods &Backbone& $mAP$ &$mAP_{50}$&$mAP_{75}$&$mAP_{s}$&$mAP_{m}$&$mAP_{l}$\\
		\hline
		Faster RCNN~\cite{ren2015faster}& ResNet-50	& 37.4  & 58.1&40.4&21.2&41.0&48.1  \\
		Faster RCNN$^+$& ResNet-50	&  39.4 & 59.4&42.5&22.6&43.3&52.2   \\
		\hdashline[3pt/ 1.5pt]
		Mask RCNN~\cite{He_2017}& ResNet-50	&  38.2 & 58.8&41.4&21.9&40.9&49.5 \\
		Mask RCNN$^+$& ResNet-50	& 39.8&59.7&43.3&22.9&43.2&53.3  \\
		\hdashline[3pt/ 1.5pt]
		Libra RCNN~\cite{pang2019libra}& ResNet-50	& 38.3 &59.5&41.9&22.1&42.0&48.5  \\
		Libra RCNN$^+$& ResNet-50	 & 40.5 &60.1&43.9&23.0&43.4&54.2   \\
		\hdashline[3pt/ 1.5pt]
		Guided Anchoring~\cite{wang2019region}& ResNeXt-101	&  43.0 & 62.5&46.9&24.6&47.4&56.1\\
		Guided Anchoring$^+$& ResNeXt-101	&43.8  &62.7 &48.0&25.1&47.7&57.8 \\	
		\hline
	\end{tabular}
	\label{tab:COCOResults}}
\end{table*}

\textbf{Simulation Experiment. }
\label{exp_se}
Label assignment takes place prior to model training and serves the purpose of selecting high-quality positive and negative samples from a vast pool of raw samples. 
The outcome of this assignment significantly influences the subsequent model training process.
We believe that the single criterion label assignment results in inadequate training samples for small-sized ground-truth, which is a significant factor contributing to poor performance in small object detection.
To prove this hypothesis, we design a simulation process for the label assignment, which emphasizes the insufficient training of small objects in current mainstream detectors and demonstrates improvement of our proposed Multiple Criteria Label Assignment~(MCLA).

Currently, most detectors~\cite{He_2017, cai2018cascade, yang2021r3det, han2021align, han2021redet, yang2019reppoints, xu2020gliding, xie2021oriented} generally employ the MaxIoU strategy for label assignment.
To maintain generality, we select the most common one-stage MaxIoU and two-stage MaxIoU, which follow the designs of RetinaNet~\cite{lin2017focal} and Faster RCNN~\cite{ren2015faster}, respectively.
Additionally, we include MCLA in the simulated assignment to facilitate comparison with other schemes.
The details of the simulation experiment are as follows.
First, we follow the original method of anchor generation to obtain initial boxes on the five predefined feature layers.
Second, we randomly generate $2,000$ ground truth boxes on an $800 \times 800$ image, with their positions, sizes, and aspect ratios uniformly distributed.
Note that the maximum dimension of ground truth not exceeding $64$ pixels. 
Third, we assign positive and negative labels using three strategies: MCLA, MaxIoU (One-stage), and MaxIoU (Two-stage). 
Finally, we count and calculate the number of positive samples by size scale after completing the assignment.

\begin{figure}[!t]
\centering
\includegraphics[width=\linewidth]{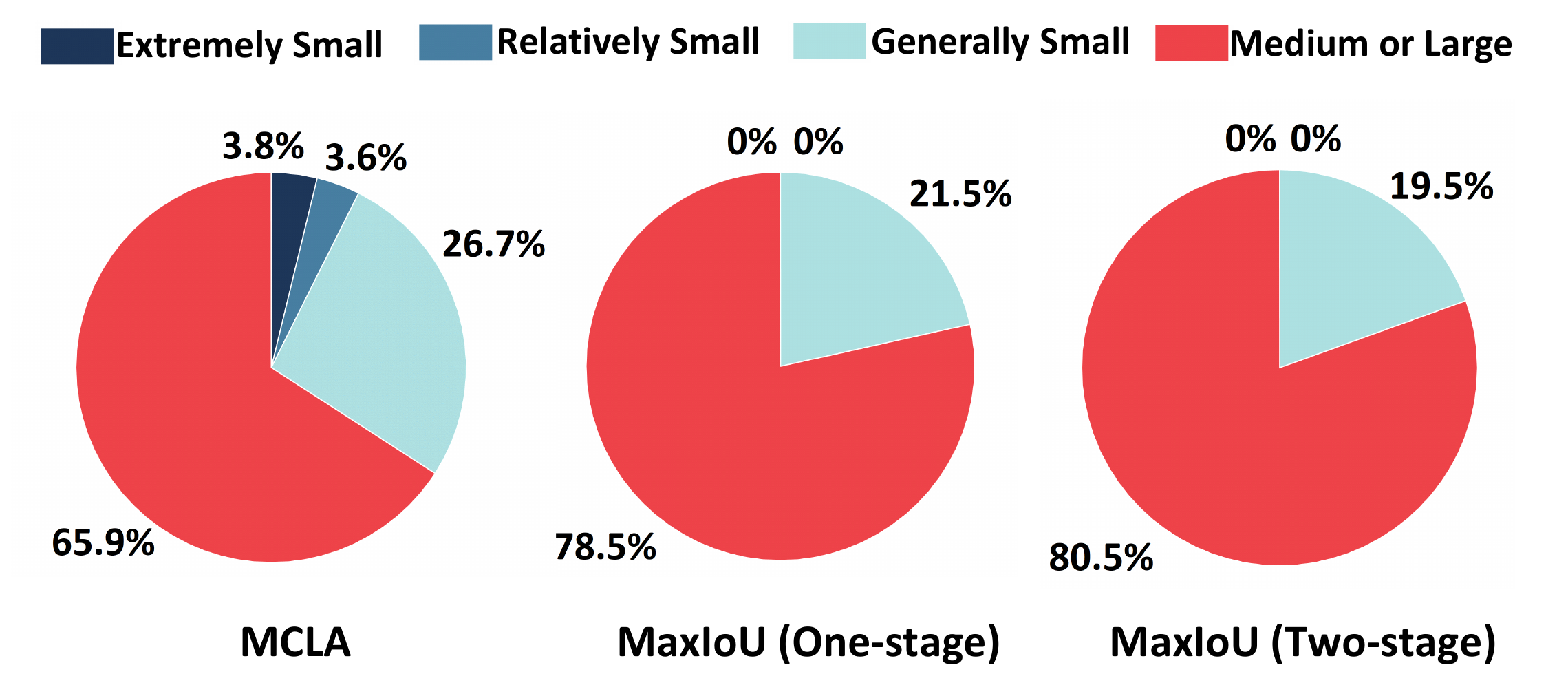}
\caption{ 
	Proportion analysis of simulation experiment. The percentages in the pie chart show the proportion of positive sample for different scales. According to the definition of the SODA-D dataset~\cite{cheng2023towards}, small objects are categorized into Extremely Small (0 to 12 pixels), Relatively Small (12 to 20 pixels), and Generally Small (20 to 32 pixels).
}
\label{abfig1}
\end{figure}

Fig.~\ref{abfig1} shows the results of simulation experiments. 
It illustrates the percentage of positive samples assigned by three different label assignment schemes across scales.
We can intuitively observe that the representative MaxIoU-based methods completely ignore the extremely small and relatively small objects, while our proposed MCLA assigns 7.4\% positive labels to these objects. 
For generally small objects, MCLA assigns 26.7\% of positive samples, which is 5.2\% and 7.2\% higher than the one-stage and two-stage MaxIoU-based methods, respectively.
The statistical results clearly demonstrate the limitations of IoU-based label assignments for small objects. 
On the contrary, our proposed MCLA method assigns positive samples in a more reasonable manner. 
In particular, it pays close attention to the extremely small and relatively small objects and assigns a few positive samples to them, thus improving the detection capability for those small objects.

\subsection{Visualizations}
\begin{figure*}[thbp]
\centering
\includegraphics[width=\linewidth]{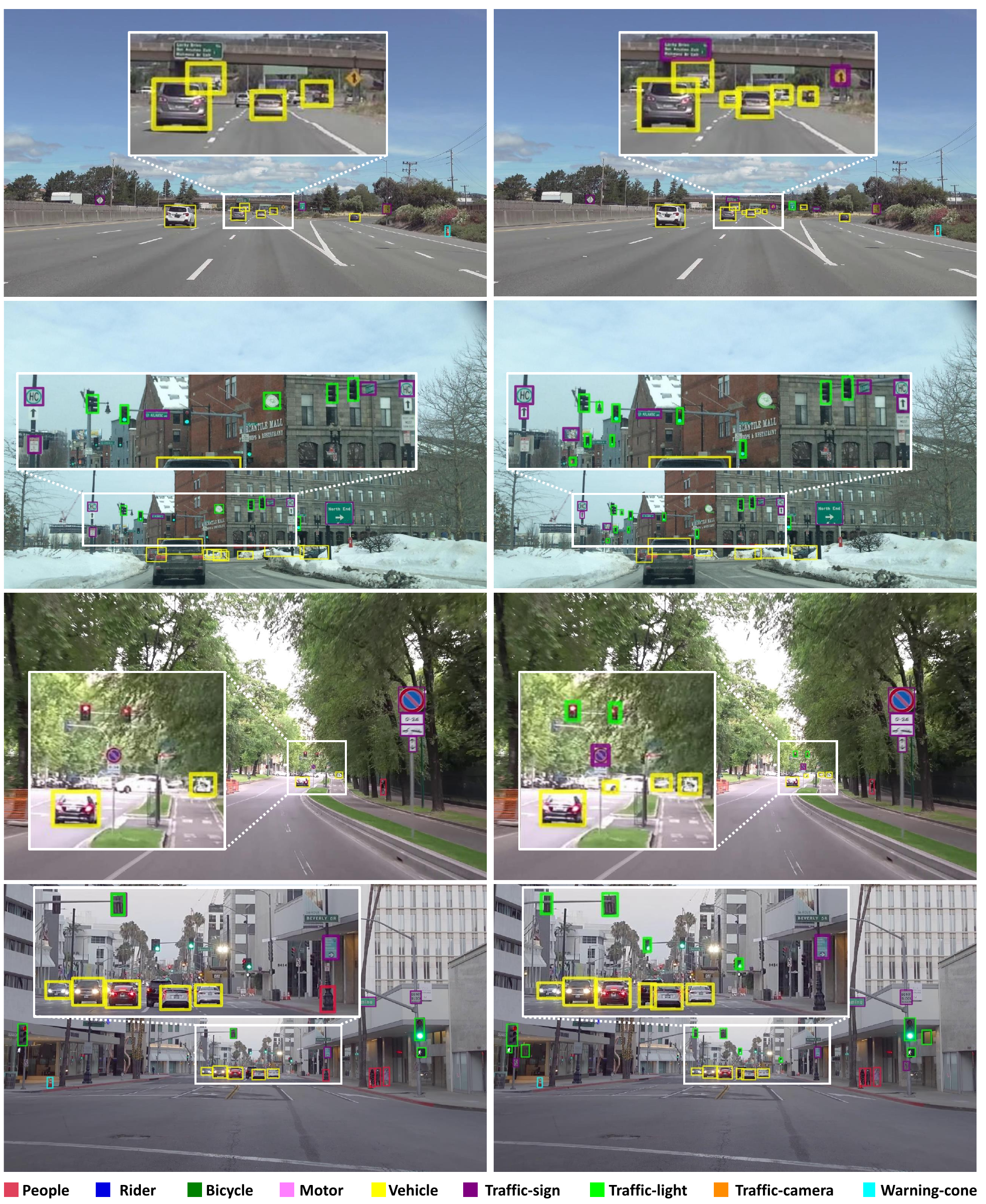}
\caption{Visual comparison between the baseline (Faster RCNN) and PLUSNet on the SODA-D \texttt{test-set}. The first column displays the visualization results of the baseline, while the second column showcases the visualization results of PLUSNet. To facilitate comparison, we have zoomed in on specific regions. It is important to note that the SODA-D dataset applies masking to confine the detection objects within a sufficiently small area. This operation directly removes the pixels of larger objects from the original images, leading to the failure of detecting larger objects, as exemplified by the larger vehicle in the second image.}
\label{vis1}
\end{figure*}

\textbf{Visualizations on SODA-D. }
\label{vis_sodad}
We present several visual results on the SODA-D dataset and compare them with the baseline method, \ie, Faster RCNN~\cite{ren2015faster}.
Through the visualizations in Fig.~\ref{vis1}, the greater potential of PLUSNet for small object detection becomes more intuitive.
For objects with small sizes and limited features, such as the blurry traffic lights, traffic signs, and vehicles, PLUSNet demonstrates superior capability in detection and accurate classification. 
This observation highlights the advantage of our approach in detecting distant small objects in the field of autonomous driving, enabling intelligent vehicles to make timely decisions.

\begin{figure*}[htbp]
\centering
\includegraphics[width=\linewidth]{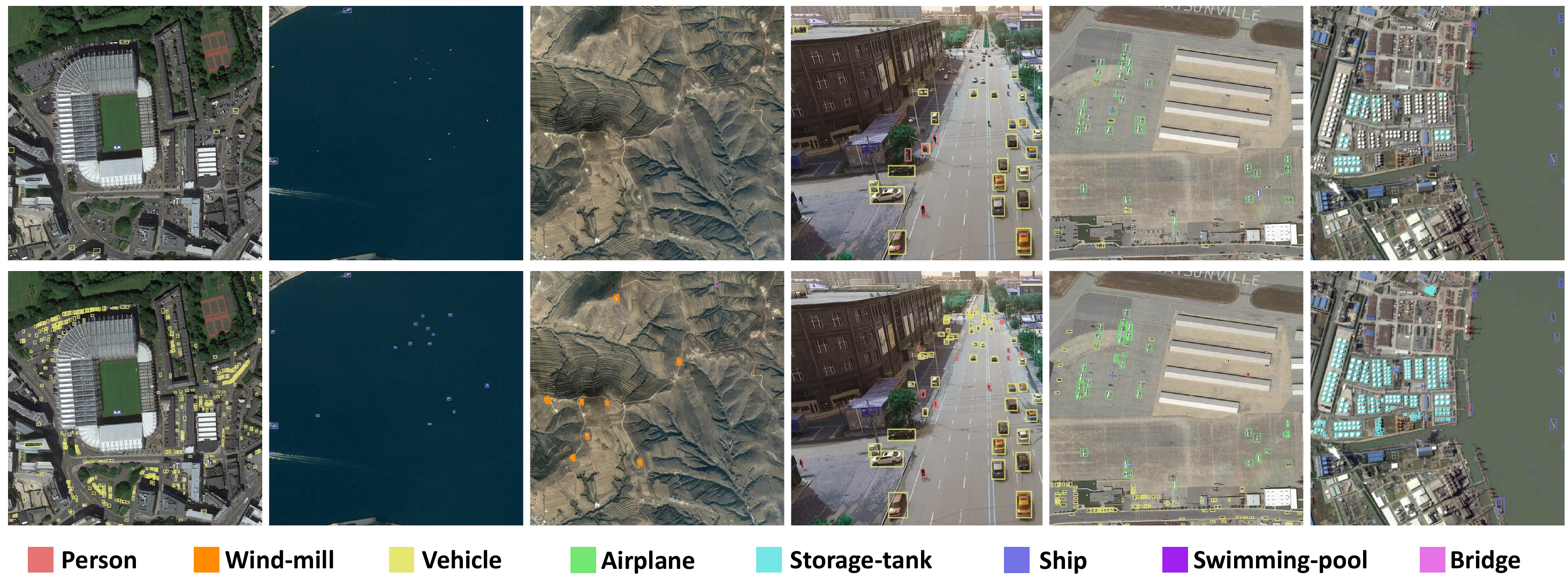}
\caption{Visual comparison between the baseline (Faster RCNN) and PLUSNet on the AI-TOD \textit{test-set}. The first row presents the visualization results of the baseline, while the second row displays the visualization results of PLUSNet. Our method significantly enhances the capability of the baseline algorithm to detect small objects. Zoom in for easier viewing.}
\label{vis2}
\end{figure*}

\textbf{Visualizations on AI-TOD. }
\label{vis_aitod}
We also conduct visualizations and comparisons with the baseline method (Faster RCNN) on the AI-TOD dataset, which focuses on small objects in remote sensing. 
The dataset primarily consists of satellite remote sensing images, with a few UAV~(unmanned aerial vehicle) perspective images. 
These images share common characteristics, including small-sized objects that are often densely distributed. 
Additionally, the overhead perspective introduces various angles, further increasing the difficulty of detection.
As depicted in the Fig.~\ref{vis2}, the baseline method achieves accurate detection results for objects of general sizes, such as larger vehicles or persons. 
However, as the object scale decreases, the baseline method struggles to detect small objects effectively, resulting in a high miss rate. 
In contrast, our proposed PLUSNet consistently demonstrates excellent detection performance across various scenarios.
Particularly, it excels in detecting small-sized objects such as tiny ships, tiny vehicles, wind-mills, storage-tanks, \etc, exhibiting the progressive ability to detect small objects of different categories. 
And the notable performance improvement is attributed to the innovative PLUS modules.
These compelling results strongly emphasize the superior performance and untapped potential of our method in the field of remote sensing object detection.

\subsection{Limitation}
\label{limit}
We must acknowledge that although utilizing frequency domain information has greatly benefited feature purification and decoupling of detection heads for small object detection, the computational cost introduced by Fourier Transform is significant. 
We compare the performance of baseline method, state-of-the-art method, and our approach on the SODA-D dataset. 
The results are shown in Tab.~\ref{tab:limit}.
While PLUSNet only exhibits a slight increase in parameter count, the computational overhead is significantly higher. 
As an exploratory work that incorporates frequency domain information into small object detection, we believe there is still ample room for optimization. 
For example, alternative and more efficient frequency domain transformation methods could be explored, or the model could be guided to learn feature patterns after Fourier Transform during the training process, enabling direct inference with these transformed features.
\begin{table*}[t!]
	\centering
	\caption{Comparison of accuracy and speed performance among baseline method, state-of-the-art method, and our approach.}
	\resizebox{1.0\textwidth}{!}{
		\begin{tabular}{|c|c|ccccc|cc|}
			\hline
			Method & Backbone & $AP$  & $AP_{eS}$  & $AP_{rS}$  & $AP_{gS}$  & $AP_{N}$ &Params & GFLOPs \\
			\hline
			Faster RCNN \cite{ren2015faster} & ResNet-50 & 28.9  & 13.8  & 25.7  & 34.5  & 43.0& \textbf{41.16 M}&\textbf{206.7} \\
			CFINet \cite{yuan2023small} & ResNet-50 & 30.7 & 14.7 & 27.8 & 36.4 & 44.6 &43.98 M &227.1\\
			PLUSNet (ours) & ResNet-50 &  \textbf{32.0} & \textbf{15.4} & \textbf{28.5} & \textbf{38.2} & \textbf{47.0} &47.63 M&481.0\\
			\hline
		\end{tabular}
	}
	\label{tab:limit}
\end{table*}%

\section{Conclusion}
In this paper, we present PLUSNet, a high-quality pipeline designed to address the specific challenges of small object detection. 
PLUSNet incorporates three plug-and-play modules: HFP, MCLA, and FDHead. 
They improve original design at the upstream, midstream, and downstream stages of the detection pipeline, resulting in a mutually beneficial outcome.
Our approach consistently achieves remarkable advancements across various datasets. 
These results underscore the importance of systematically enhancing the entire detection pipeline.
Notably, this is the first attempt to scrutinize object detection from a holistic perspective and employ relatively naive components for improvement. 
There is still ample room for exploration in terms of how to ensure the synergy among various components of the pipeline and further optimize their performance.


\newpage
\bibliographystyle{IEEEtran}
\bibliography{refbib}

\end{document}